%% file: main.tex
\documentclass{article} 
\usepackage[accepted]{icml2019}
\usepackage{graphicx}
\usepackage{amsmath,amssymb}
\usepackage{natbib}
\usepackage{makecell}
\usepackage{tikz}

\usepackage{hyperref}
\usepackage{url}

\usepackage{pifont}
\newcommand{\cmark}{\ding{51}}%
\newcommand{\xmark}{\ding{55}}%

\icmltitlerunning{Unsupervised learning of influential trajectories}

\begin{document}

\twocolumn[
\icmltitle{The Journey is the Reward: \\
    Unsupervised Learning of Influential Trajectories}
    
\begin{icmlauthorlist}
\icmlauthor{Jonathan Binas}{mila}
\icmlauthor{Sherjil Ozair}{mila}
\icmlauthor{Yoshua Bengio}{mila}
\end{icmlauthorlist}
\vspace{2em}

\icmlaffiliation{mila}{Mila, Montreal}

\icmlcorrespondingauthor{Jonathan Binas}{jbinas@gmail.com}

]

\printAffiliationsAndNotice{}

\begin{abstract}
Unsupervised exploration and representation learning become increasingly important when learning in diverse and sparse environments.
The information-theoretic principle of empowerment formalizes an unsupervised exploration objective through an agent trying to maximize its influence on the future states of its environment.
Previous approaches carry certain limitations in that they either do not employ closed-loop feedback or do not have an internal state. As a consequence, a privileged final state is taken as an influence measure, rather than the full trajectory.
We provide a model-free method which takes into account the whole trajectory while still offering the benefits of option-based approaches. We successfully apply our approach to settings with large action spaces, where discovery of meaningful action sequences is particularly difficult.
\end{abstract}

\section{Introduction}

Efficient exploration and representation learning are two core challenges in reinforcement learning.
An agent that understands how its environment works, in particular the causal structure of the environment, knows the consequences of its behavior and will be able to learn new tasks quickly.
Such structural knowledge is not necessary linked to a task and ideally is acquired before a specific task is learned.

An information-theoretic principle known as \emph{empowerment} has led to a number of approaches for task-agnostic exploration and representation learning (e.g. \citet{mohamed2015variational,gregor2016variational}).
Empowerment formulates an unsupervised objective which can be thought of as finding an optimal code for transmitting information through the environment. Considering the environment an information channel and maximizing the mutual information between actions and effects means to understand how to control the environment in such a way that particular states can be achieved.

Another motivation for using such information-based objectives in reinforcement learning is that utilizing an information channel in an optimal way can be linked to the emergence of compact representations \citep{hoel2017map}.
Finding an optimal code translates to finding an optimal action distribution which influences the environment most effectively.
While finding this distribution is manageable in simple environments, the search space becomes intractable and the problem notoriously hard in settings with large action spaces, where actions need to be coordinated to achieve something meaningful.

Previous approaches have not tackled such problems, where a meaningful representation of the action space needs to be learned, and thus potentially do not leverage the full potential of this method.
Moreover, to our knowledge, no other approach is model free, suited for partially observable settings, and uses closed-loop feedback.
We introduce a model-free approach with memory and closed-loop feedback, such that control over full trajectories can be optimized, rather than just control over a desired final state of a trajectory.
We demonstrate that our model learns to understand a complex environment without external reward or supervision.

\section{Empowerment -- a brief review}

Empowerment is a popular objective for task-agnostic reinforcement learning. \citet{klyubin2005empowerment} define \emph{empowerment} as the maximum mutual information in the agent-to-environment information channel, as a measure of how much control an agent exerts on its environment. They argue that an agent should strive to increase its empowerment in the absence of any specific tasks. For an extensive introduction, please refer to \citet{salge2014empowerment}.

\citet{jung2011empowerment} conduct an extensive set of experiments to show its applicability. In particular, they show that empowerment maximization can swing up a pendulum without any rewards for swinging up.

\citet{mohamed2015variational} maximize a lower bound to the mutual information and learn open-loop options using deep neural networks on a variety of nontrivial gridworld environments.

\citet{gregor2016variational} extend it to closed-loop options and show that closed-loop options achieve higher empowerment. They also propose an option-less empowerment-maximizing policy which is able to deeply explore a first-person 3D environment with pixel observations.

\citet{tiomkin2017unified} show that empowerment admits a Bellman-like recursive formulation, and thus can be optimized using a temporal difference version of the Blahut-Arimoto algorithm. However, their formulation does not admit a way to learn options (similar to \citet{gregor2016variational}'s second algorithm).

\citet{eysenbach2018} build on top of \citet{gregor2016variational}'s first algorithm with maximum-entropy policies and a fixed prior over options, which enables stable learning demonstrated on continuous control tasks.

\citet{thomas2018disentangling} present an approach, where options are explicitly mapped to variations of the state.

All the above methods except \citet{tiomkin2017unified} consider final states or observations as a proxy for behavior. Our work learns options for trajectories, i.e. two distinct sequences of states are considered different even if they share the same final state. This makes our proposed variant of empowerment particularly suitable for partially-observable environments. 
Table~\ref{table:empowerment} presents a comparison of these differing variants of empowerment, including our own proposal.

\begin{table}
\begin{center}
\small
\begin{tabular}{l|c|c|c}
    \thead{Method} & \thead{Closed \\ loop opt.} & \thead{Partial \\ obs.} & \thead{Model-\\free} \\
    \hline
    \citet{klyubin2005empowerment} & \xmark & \xmark & \xmark \\
    \citet{jung2011empowerment} & \xmark & \xmark & \xmark \\
    \citet{mohamed2015variational} & \xmark & \cmark & \cmark\\
    \citet{gregor2016variational} Alg. 1 & \cmark & \xmark & \cmark\\
    \citet{gregor2016variational} Alg. 2 & - & \cmark & \cmark \\
    \citet{tiomkin2017unified} & - & \xmark & \xmark\\
    \citet{thomas2018disentangling} & \cmark & \xmark & \cmark \\
    \citet{eysenbach2018} & \cmark & \xmark & \cmark\\
    \textbf{This work} & \cmark & \cmark & \cmark \\
    \hline
\end{tabular}
\end{center}
\caption{A comparison of various empowerment variants proposed in the literature. Dashes indicate that the corresponding variant does not learn options.}
\label{table:empowerment}
\end{table}

\section{Model}

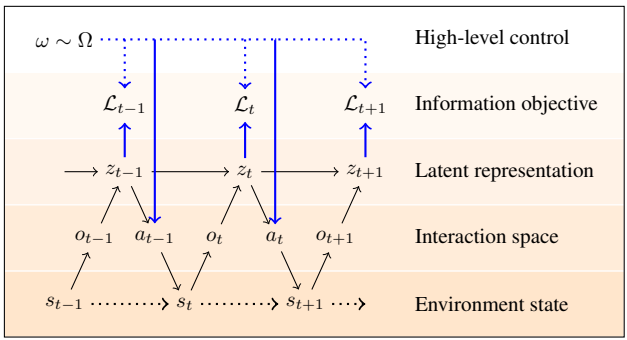
\begin{figure}
    \centering
    \input{figures/model.tex}
    \caption{Illustration of the model. Blue arrows correspond to the parts that are added on top of a regular agent model with memory. Here, $\omega$ represents an option, $\mathcal{L}$ the mutual information objective, $z_t=f(o_t, z_{t-1})$ the latent state, $a_t\sim\pi(a_t|z_t,\omega)$ an action, $o_t=o(s_t)$ an observation, and $s_t$ the state. In our implementations, we do not optimize $\mathcal{L}$ directly, but rather provide a reward signal $r_t = \delta_{\omega,\omega'_t}$, where $\omega'_t \sim q(\omega|z_t)$ is the option inferred at time $t$. Solid lines are learned.}
    \label{fig:model}
\end{figure}

We consider a (partially observable) Markov decision process (MDP), defined by $(S, A, \Gamma, R)$, where $S$ is a finite set of states, $A$ is the set of actions, $\Gamma = p(s_{t+1} | s_t, a_t)$ is the state transition function, and $R_a(s, s')$ is an extrinsic reward function, specifying the reward received when transitioning from state $s$ to $s'$ due to action $a$.
At every timestep $t$, the agent receives an observation $o_t$, where $o_t = o(s_t)$, and emits an action $a_t \in A$. In our unsupervised setting, we assume the external reward $r_t = 0$ for all $t$.
The self-supervised agent model is defined through an information source $\Omega$, a latent state $z_t = f(o_t, z_{t-1})$, a policy $\pi(a_t | z_t, \omega_t)$, where $\omega_t$ corresponds to a sample from $\Omega$ and is to be encoded in the agent's actions, and an inverse model $q(\omega | z_t)$.
Our objective for unsupervised exploration and representation learning boils down to maximizing the mutual information between the information source $\Omega$ and a representation of a sequence of observations, $z_t = f(o_t, z_{t-1})$. Thereby, rather than using the latent state distribution $p(z_t | o_t, z_{t-1})$ directly, we infer the original information using a learned function $q$, and thus,
\begin{align}
    \max_{\pi, q, f} \hat I(\Omega; q(\omega | z_t))\,,
    \label{mi1}
\end{align}
where we train $\pi$, $q$, and $f$ simultaneously. The approximation in eqn.~\ref{mi1} corresponds to a variational lower bound of $I(\Omega; \{o_0,\ldots,o_T\})$, assuming the agent was acting over $T$ timesteps (see Appendix~\ref{sect:lower} for details).
The latent state enables our model to maximize information transmitted into a trajectory rather than a single (final) state.
In our implementations, rather than providing an internal reward at every timestep, we choose $\Omega$ to be a uniform distribution over a discrete space, and we sample $\omega_t \sim q$ and provide an internal reward of 1 whenever $\omega_t$ matches the original input word sampled from $\Omega$.
With this reward, our model can be optimized using any reinforcement learning algorithm.
The model is illustrated in fig.~\ref{fig:model}.

\begin{figure*}[t]
    \centering
    \includegraphics[width=\textwidth, trim=5cm 3cm 4cm 0cm]{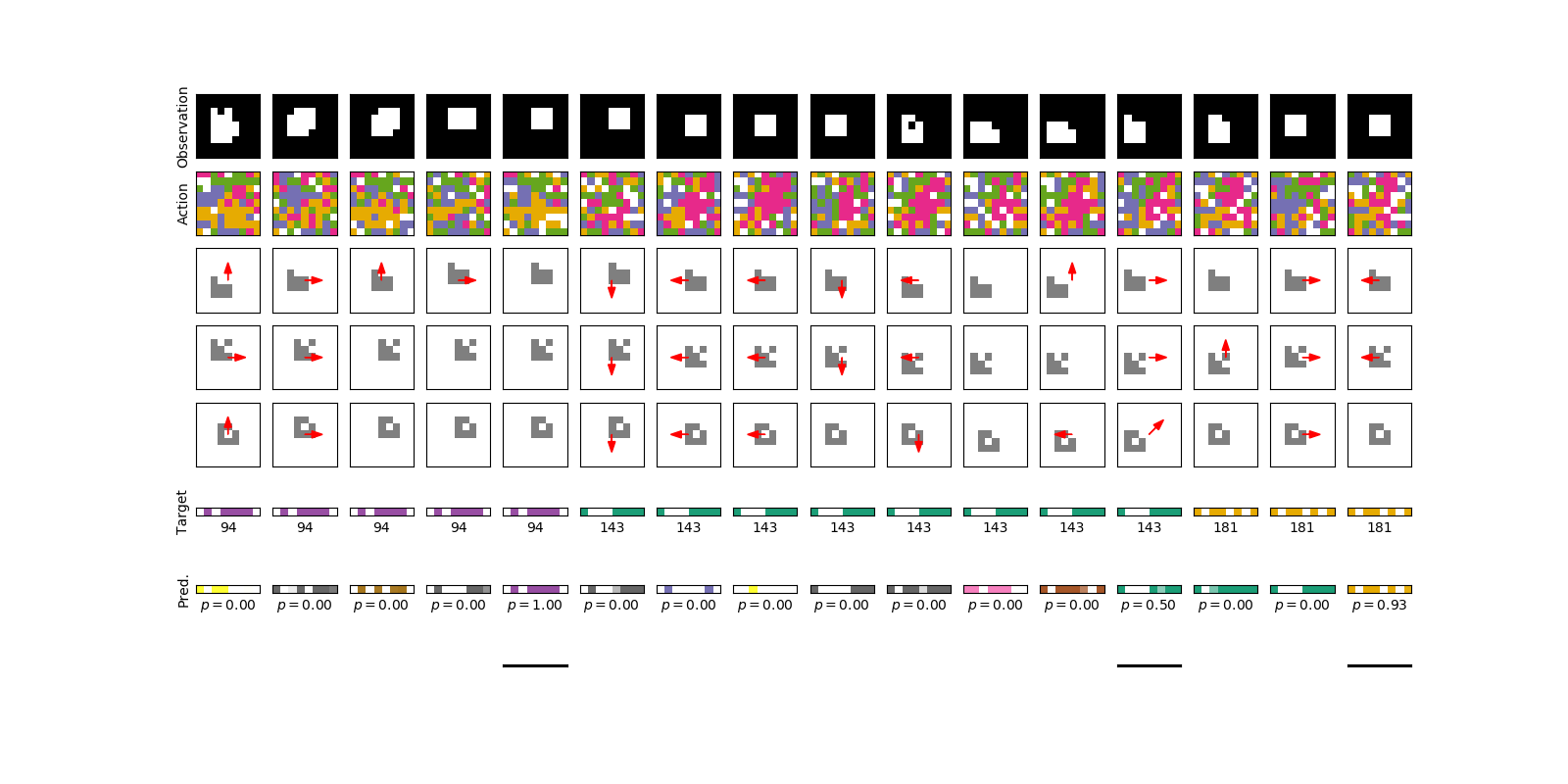}
    \caption{Unsupervised learning of control. 1) The environment consists of random patches of size $3\times3$, which are potentially overlapping, and which can be moved around by exerting forces on individual pixels. A patch will move towards a certain direction if the net force applied to it exceeds a threshold. 2) For each pixel $i$, the agent emits an action (force) $f_i \in \{\leftarrow, \downarrow, \rightarrow, \uparrow, \circ\},$ where $\circ$ represents no force. The action space is thus of size $5^n$, where $n$ is the number of pixels in the environment (81 in this case). Different actions are represented in different colors; white represents no action. 3) The middle panels show the internal state and dynamics of the environment. Red arrows represent the forces resulting from the action. 4) A random target word (or option) to be transmitted through the environment is provided to the agent; a new word is drawn whenever the agent infers the provided word correctly (indicated at the bottom). 5) Probabilities of the predicted bits being 1, and probability of the target under the inverse distribution.}
    \label{fig:res1}
\end{figure*}

\subsection{Derivation of the lower bound}
\label{sect:lower}

We intend to optimize a lower bound of
\begin{align}
    I(\Omega; p(o_0, \ldots, o_T))\,,
\end{align}
where $\Omega$ is the option distribution, which we assume to be uniform here, and the agent is assumed to interact with the environment over $T$ timesteps.
We have
\begin{align*}
    z_T &= f(z_{T-1}, o_T) \\
        &= f(f(z_{T-2}, o_{T-1}), o_T) = \ldots =: F(o_0, \ldots, o_T)\,,
\end{align*}
and thus we consider $z_T$ an embedding of the full observation history. With the causal structure of the model and the data processing inequality we get
\begin{align*}
     I(\Omega, p(o_0, \ldots, o_T)) &\geq I(\Omega; p(z_T)) \\
        &= \left\langle \log \frac{p(\omega | z_T)}{p(\omega)} \right\rangle_{p(z_T|\omega)p(\omega)}\,.
\end{align*}
The right-hand side can further be written as
\begin{align*}
    I(\Omega; p(z_T)) = \left\langle \log p(\omega | z_T) \right\rangle_{\omega,z_T} + H(\Omega)\,,
\end{align*}
where $H(\Omega)$ is a constant and therefore ignored in the optimization.
We can now approximate the conditional with a variational distribution, and simply optimize
\begin{align}
    \max \hat I(\Omega; p(z_T)) = \max \left\langle \log q(\omega | z_T) \right\rangle_{\omega,z_T}\,,
\end{align}
which we call the \emph{empowerment} objective. We optimize this objective over the policy parameters using reinforcement learning.

\section{Experiments}

\paragraph{Pushing boxes.} To demonstrate the effectiveness of our approach, we evaluate its performance on a synthetic task where the agent receives a top-down view of an environment containing several objects (random patches of $3\times3$ pixels). Patches can overlap and thus the boundaries between objects might not be visible.
The agent can exert a force on each individual pixel, which can be one of up, down, left, right. It can also choose to apply no force. The forces of individual pixels are transferred to the respective objects and if the net force applied to an object exceeds some threshold (i.e. if the forces are sufficiently aligned) the object moves into the respective direction by a small distance.
We consider an environment of size $9\times9$ and three objects, which are initialized at random locations inside the field of view.
This environment is particularly challenging because of its large action space and the fact that objects can occlude each other.
Initially, an option $\omega$ is drawn from $\Omega$, and is represented as a binary string of a certain length (e.g. 8 bits in the example shown in fig.~\ref{fig:res1}.)
The agent rewards itself as soon as it correctly infers the original string $\omega$ based on its latent state $z$, and subsequently draws a new option which it then tries to encode in its trajectory.
In the case of 8 bits, random guessing would lead to a correct guess every 256 steps on average. The agent learns, however, to encode information in its actions in such a way that they influence the environment state sufficiently for the agent to infer the option from its observations after only a small number of steps (less than 10).
It is to be noted that in this environment random policies emitting uncoordinated actions typically have no effect at all, since only coordinated (aligned) actions can exceed the threshold for shifting a block.
For similar reasons, the agent learns to never push the blocks outside of the field of view (there are no walls) since, as soon as not enough pixels of a block are visible, the actions applied to these pixels are not sufficient to move the block and it becomes useless.

Notably, as can be seen in fig.~\ref{fig:res1}, the agent causes the environment to produce the same observation in different contexts (all patches centered in the image; step 8 and step 16). However, the trajectories leading up to those states are different, and thus information is still transmitted. This scenario could not be solved by simply using the current state.

As a performance metric for the proposed approach, we consider the median number of steps the agent requires to recover the option $\omega$. A naive guesser would, on average, require a number of steps of the order of the number of available options, $|\Omega|$, for a uniform option distribution. The trained model performs substantially better than the baseline agent.

\begin{table}[h]
    \centering
    \begin{tabular}{l|cc}
        \thead{Model} & \thead{\#options} & \thead{\#steps} \\
        \hline
         Baseline & 16 & 11.0 \\
         Empowered & 16 & \bf 3.0 \\
         Baseline & 256 & 179.0 \\
         Empowered & 256 & \bf 9.0 
    \end{tabular}
    \caption{Empowerment performance of the model trained with the trajectory-based objective. Numbers correspond to the median over a minibatch of trials.}
    \label{tab:perf}
\end{table}

The fact that the agent is able to transmit information through the environment means that it is able to reliably choose and decode a large number of trajectories, which it discovers through unsupervised training.


\paragraph{Training and model details.}

The agent model consisted of a 3-layer convnet, generating an image embedding, an LSTM with 256 hidden units as a memory model, a single layer, fully connected network for the critic, and a 3 layer, fully connected network for the policy. The inverse model, $q$, was implemented using a single fully-connected layer as well. ReLU activations were used between layers and sigmoid activations were used on the outputs of $q$ and $\pi$ to generate probability values.

All models were trained using the proximal policy optimization algorithm \citep{schulman2017proximal}. The hyperparameters used are listed in Appendix~\ref{sect:hyper}.

\section{Conclusion}

We propose a new variant of an empowerment-based, unsupervised learning method, which is suitable for partially observable settings, model-free, and contains a closed-loop option mechanism.
Unlike other option-based methods, which privilege the final state of a trajectory, our approach uses the full trajectory to infer effects of the agents actions. Thus, the intrinsic reward is based on the whole journey, not just the goal.
We successfully train an agent to control a complex environment, purely based on intrinsic reward signals. We focus on a task featuring high-dimensional actions, which make the problem harder but also more interesting, as for successful acting, good representations have to be learned for both observations and actions.
Future work will include combining the unsupervised system with more complex RL tasks and more detailed analyses of more complex environments.

\subsubsection*{Acknowledgments}
We thank Anirudh Goyal, Junhao Wang, and our colleagues at Mila for helpful discussions.

\vfill

\bibliography{bibliography}
\bibliographystyle{humannat}

\vfill\eject

\appendix

\section{Model parameters}
\label{sect:hyper}

\begin{table}[ht]
    \centering
    \begin{tabular}{lc}
        Batch size & 256 \\
        Clipping $\epsilon$ & 0.2 \\
        Discount factor & 0.99 \\
        GAE $\lambda$ & 0.95 \\
        Learning rate & 0.0007 \\
        Max grad norm & 0.5 \\
        Adam $\epsilon$ & 1e-05 \\
        Processes & 64 \\
        Recurrence (16 options) & 4 \\
        Recurrence (256 options) & 8 \\
        Value loss term coeff. & 0.5
    \end{tabular}
    \caption{PPO hyperparameters used in experiments.}
    \label{tab:params}
\end{table}

\end{document}

%% file: figures/model.tex
\begin{tikzpicture}[scale=0.8, every node/.append style={scale=0.8}, yscale=1.1]

    \draw[thin, white, fill=orange!20] (-3,-0.5) rectangle (7.3,0.5);
    \draw[thin, white, fill=orange!15] (-3,0.5) rectangle (7.3,1.5);
    \draw[thin, white, fill=orange!10] (-3,1.5) rectangle (7.3,2.5);
    \draw[thin, white, fill=orange!5] (-3,2.5) rectangle (7.3,3.5);
    \draw (-3,-0.5) rectangle (7.3,4.5);
    
    \node (x0) at (-2,0) {$s_{t-1}$};
    \node (x1) at (0,0) {$s_t$};
    \node (x2) at (2,0) {$s_{t+1}$};
    
    \node (z0) at (-1,2) {$z_{t-1}$};
    \node (z1) at (1,2) {$z_t$};
    \node (z2) at (3,2) {$z_{t+1}$};
    
    \node (o0) at (-1.5,1) {$o_{t-1}$};
    \node (o1) at (0.5,1) {$o_{t}$};
    \node (o2) at (2.5,1) {$o_{t+1}$};
    \node (a0) at (-0.5,1) {$a_{t-1}$};
    \node (a1) at (1.5,1) {$a_{t}$};
    
    \node (w0) at (-2,4) {$\omega\sim\Omega$};
    \node (l0) at (-1,3) {$\mathcal{L}_{t-1}$};
    \node (l1) at (1,3) {$\mathcal{L}_{t}$};
    \node (l2) at (3,3) {$\mathcal{L}_{t+1}$};
    
    \draw[thick, dotted, ->] (x0) -- (x1);
    \draw[thick, dotted, ->] (x1) -- (x2);
    \draw[thick, dotted, ->] (x2) -- ++(1,0);
    \draw[<-] (z0) -- ++(-1,0);
    \draw[->] (z0) -- (z1);
    \draw[->] (z1) -- (z2);
    \draw[->] (z0) -- (a0);
    \draw[->] (a0) -- (x1);
    \draw[->] (z1) -- (a1);
    \draw[->] (a1) -- (x2);
    \draw[->] (x0) -- (o0);
    \draw[->] (x1) -- (o1);
    \draw[->] (x2) -- (o2);
    \draw[->] (o0) -- (z0);
    \draw[->] (o1) -- (z1);
    \draw[->] (o2) -- (z2);
    \draw[thick, ->, blue] (w0-|a0) -| (a0);
    \draw[thick, ->, blue] (w0-|a1) -| (a1);
    \draw[thick, ->, blue, dotted] (w0) -| (l2);
    \draw[thick, ->, blue, dotted] (w0-|l0) -- (l0);
    \draw[thick, ->, blue, dotted] (w0-|l1) -- (l1);
    \draw[thick, ->, blue] (z0) -- (l0);
    \draw[thick, ->, blue] (z1) -- (l1);
    \draw[thick, ->, blue] (z2) -- (l2);

    \node[right] at (3.7,0) {Environment state};
    \node[right] at (3.7,1) {Interaction space};
    \node[right] at (3.7,2) {Latent representation};
    \node[right] at (3.7,3) {Information objective};
    \node[right] at (3.7,4) {High-level control};
    
\end{tikzpicture}

%% file: main.bbl
\begin{thebibliography}{}

\bibitem[\protect\astroncite{Eysenbach et~al.}{2018}]{eysenbach2018}
Eysenbach, B., A.~Gupta, J.~Ibarz, and S.~Levine\leavevmode\nopagebreak\newline
  2018.
\newblock Diversity is all you need: Learning skills without a reward function.
\newblock {\em arXiv preprint arXiv:1802.06070}.

\bibitem[\protect\astroncite{Gregor et~al.}{2016}]{gregor2016variational}
Gregor, K., D.~J. Rezende, and D.~Wierstra\leavevmode\nopagebreak\newline 2016.
\newblock Variational intrinsic control.
\newblock {\em arXiv preprint arXiv:1611.07507}.

\bibitem[\protect\astroncite{Hoel}{2017}]{hoel2017map}
Hoel, E.\leavevmode\nopagebreak\newline 2017.
\newblock When the map is better than the territory.
\newblock {\em Entropy}, 19(5):188.

\bibitem[\protect\astroncite{Jung et~al.}{2011}]{jung2011empowerment}
Jung, T., D.~Polani, and P.~Stone\leavevmode\nopagebreak\newline 2011.
\newblock Empowerment for continuous agent-environment systems.
\newblock {\em Adaptive Behavior}, 19(1):16--39.

\bibitem[\protect\astroncite{Klyubin et~al.}{2005}]{klyubin2005empowerment}
Klyubin, A.~S., D.~Polani, and C.~L. Nehaniv\leavevmode\nopagebreak\newline
  2005.
\newblock Empowerment: A universal agent-centric measure of control.
\newblock In {\em 2005 IEEE Congress on Evolutionary Computation}, volume~1,
  Pp.~ 128--135. IEEE.

\bibitem[\protect\astroncite{Mohamed and
  Rezende}{2015}]{mohamed2015variational}
Mohamed, S. and D.~J. Rezende\leavevmode\nopagebreak\newline 2015.
\newblock Variational information maximisation for intrinsically motivated
  reinforcement learning.
\newblock In {\em Advances in neural information processing systems}, Pp.~
  2125--2133.

\bibitem[\protect\astroncite{Salge et~al.}{2014}]{salge2014empowerment}
Salge, C., C.~Glackin, and D.~Polani\leavevmode\nopagebreak\newline 2014.
\newblock Empowerment--an introduction.
\newblock In {\em Guided Self-Organization: Inception}, Pp.~ 67--114.
\newblock Springer.

\bibitem[\protect\astroncite{Schulman et~al.}{2017}]{schulman2017proximal}
Schulman, J., F.~Wolski, P.~Dhariwal, A.~Radford, and
  O.~Klimov\leavevmode\nopagebreak\newline 2017.
\newblock Proximal policy optimization algorithms.
\newblock {\em arXiv preprint arXiv:1707.06347}.

\bibitem[\protect\astroncite{Thomas et~al.}{2018}]{thomas2018disentangling}
Thomas, V., E.~Bengio, W.~Fedus, J.~Pondard, P.~Beaudoin, H.~Larochelle,
  J.~Pineau, D.~Precup, and Y.~Bengio\leavevmode\nopagebreak\newline 2018.
\newblock Disentangling the independently controllable factors of variation by
  interacting with the world.
\newblock {\em arXiv preprint arXiv:1802.09484}.

\bibitem[\protect\astroncite{Tiomkin and Tishby}{2017}]{tiomkin2017unified}
Tiomkin, S. and N.~Tishby\leavevmode\nopagebreak\newline 2017.
\newblock A unified bellman equation for causal information and value in markov
  decision processes.
\newblock {\em arXiv preprint arXiv:1703.01585}.

\end{thebibliography}
